\def\year{2020}\relax
\documentclass[letterpaper]{article} 
\usepackage{aaai20}  
\usepackage{times}  
\usepackage{helvet} 
\usepackage{courier}  
\usepackage[hyphens]{url}  
\usepackage{graphicx} 
\usepackage{amsmath}

\usepackage{graphicx}  
\title{Multi-Label Product Categorization Using Multi-Modal Fusion Models }
\author{Pasawee Wirojwatanakul\textsuperscript{\rm 1} and Artit Wangperawong\textsuperscript{\rm 2}\\ 
\textsuperscript{\rm 1}New York University, pw1103@nyu.edu\\
\textsuperscript{\rm 2}U.S. Bank, 
artit.wangperawong@usbank.com
}
 
\begin{document}

\maketitle

\begin{abstract}
  In this study, we investigated multi-modal approaches using images, descriptions, and titles to categorize e-commerce products on Amazon. Specifically, we examined late fusion models, where the modalities are fused at the decision level. Products were each assigned multiple labels, and the hierarchy in the labels were flattened and filtered. For our individual baseline models, we modified a CNN architecture to classify the description and title, and then modified Keras' ResNet-50 to classify the images, achieving $F_1$ scores of $77.0\%$, $82.7\%$, and $61.0\%$, respectively. In comparison, our  tri-modal late fusion model can classify products more effectively than single modal models can, improving the $F_1$ score to $88.2\%$. Each modality complemented the shortcomings of the other modalities, demonstrating that increasing the number of modalities can be an effective method for improving the performance of multi-label classification problems. 
\end{abstract}

\noindent 
\section{Introduction}
\subsection{Background}
To sell products on many e-commerce systems, sellers are tasked with providing categories for their products. Automating product classification can reduce manual labor time and cost, giving sellers a better experience when uploading new products. Such auto-labeling can also benefit the buyers, as sellers manually tagging their own products may be inaccurate or sub-optimal. An performant classifier is important, as mislabelled products may lead to missed sales opportunities due to buyers not being able to effectively locate the things they want to buy.

Prior studies have approached product categorization as a text-classification task \cite{kozareva,yu,vandic}. However, ideally multiple types of inputs can be considered, including title, description, image, audio, video, item-to-item relationships, and other metadata. Although a few recent studies have explored product categorization using both text and images \cite{aberg,tom}, here we report on a strategy for combining an arbitrary number of inputs and modes. We specifically demonstrate a multi-modal model based on images, titles, and descriptions. Our task is different from a regular multi-class classification problem, as a product may be labeled with more than one class. Most products will appear in many classes and sub-classes. Classes can be nested in another class and potentially nested in another sub-class. Given the large amount of products uploaded and the numerous possible labels applicable, machine learning can be used to automatically classify the products in a more efficient manner.

\begin{figure*}[!hbt]
    \centering
    \includegraphics[width=0.8\textwidth]{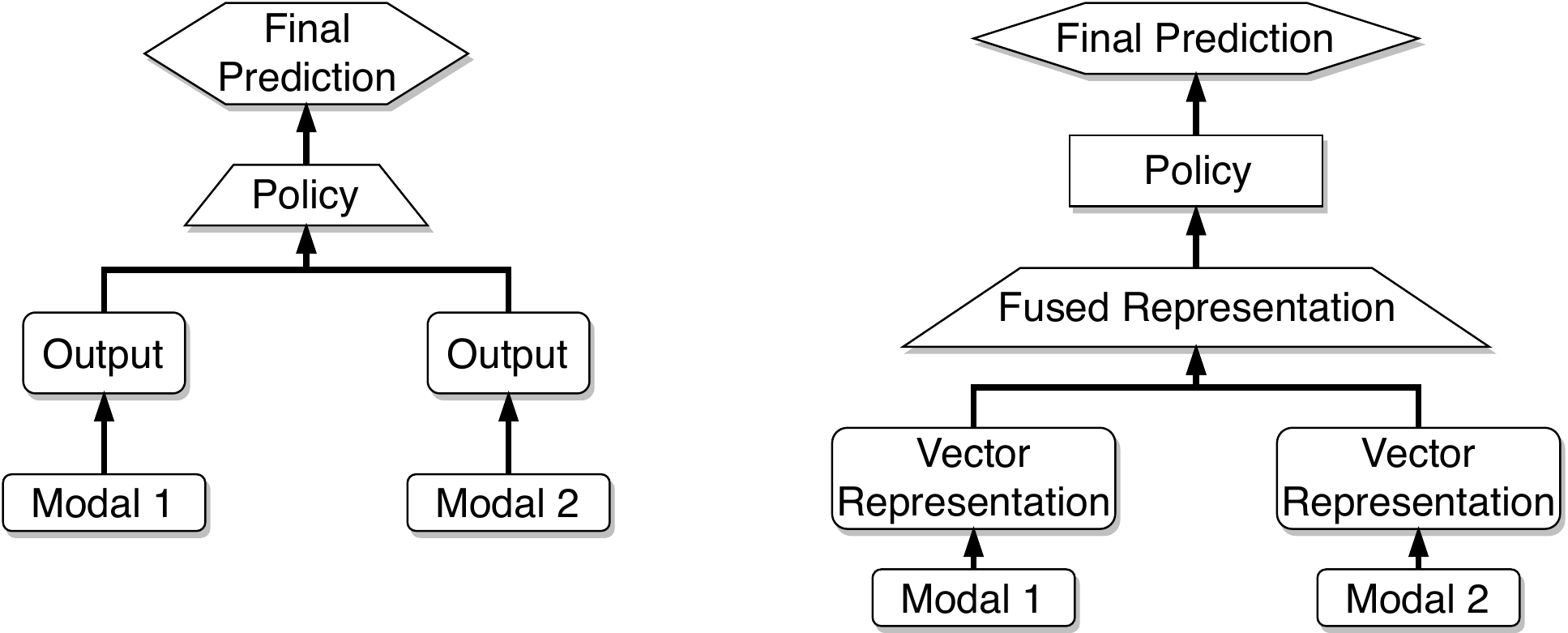}
    \caption{The diagram on the left represents late fusion and the diagram on the right represents early fusion.}
    \label{fig:1}
\end{figure*}

As images, titles and descriptions are different modalities of data that can each capture unique aspects of a product, we explored fusing individual models trained for each modality. Note that although fusing and ensembling both involve combining multiple models, for the purposes of our discussion each of the models utilize different modalities of data in fusion, whereas the same data passes through all of the models in an ensemble. There are two common ways to fuse different modal networks: late fusion and early fusion (Fig.~\ref{fig:1}). Late fusion refers to combining the predictions (outcome probabilities) of multiple networks using a certain policy. Such a policy can be using the maximum or minimum of the outcomes. In contrast, in early fusion vector representations of each modality can be extracted at an early level and fused with one another through concatenation or addition to produce a multi-modal representation vector. The model then performs classification on the resulting multi-modal representation vector.

\subsection{Related Works}
Convolutional neural network (CNN) architectures \cite{kim} have been used to classify the title of products \cite{tom}. In such a model, the first layer uses a random word embedding. The same study also used a VGG network for image classification \cite{simonyan2014deep}. While they experimented with both early and late fusion, only the late fusion resulted in an improvement in performance. The image and text classifiers were trained separately to achieve maximal performance individually before being combined by a policy network. The policy network which achieved the highest performance was a neural network with 2 fully-connected layers and took in the top-3 class probabilities from the image and text CNNs as input. Their dataset contained 1.2 million images and 2,890 possible shelves. On average, each product falls in 3 shelves. Their model is considered effective when the network correctly outputs one of the three shelves.

Existing studies have also used the image, title, and description of an ad/product to classify products into single categories \cite{aberg}. {\AA}berg concatenated the title and description, and used fastText \cite{armand} as the baseline model for text classification, while using the Inception V3 for image classification. {\AA}berg also explored a similar implementation of Kim's CNN architecture \cite{kim} but could not achieve the level of performance of fastText.

The {\AA}berg study used a dataset containing 96,806 products belonging to 193 different classes. Since each product was assigned exclusively to one class, multi-label categorization was not addressed. Hence a softmax function could be applied in the final layer before outputting the class probabilities. Similar to work of Zahavy et al. described above, both late and early fusion were explored, and late fusion yielded better results. Both heuristic policies and network policies were explored. Heuristic policies refer to some static rule; as an example, the mean of the probabilities from different modals. Network policies refer to training a neural network that takes the output probabilities from different networks and produces a new probability vector.

\section{Dataset}
The dataset used in our study comprises of 9.4 million Amazon products \cite{mcauley}. The class hierarchical information was not available, as the classes and subclasses were pre-flattened as given. We randomly sampled 119,073 products from this dataset, in which the first 90,000 products are kept for the training set. After pre-processing, there are 122 possible classes in which a product can belong to. Unlike many previous studies, here each product can be assigned multiple labels. Each product in the dataset contains the image, description, title, price, and co-purchasing network.

Product categorization systems can be challenging to build due to the trade-off between the number of classes and efficacy. As an example, adding more classes and sub-classes to a product might make it easier to discover, but more classes would also increase the likelihood of an incorrect class being applied. To address this issue, some studies reduced the number of sub-classes \cite{tom,stanford}. One method is to create a shelf and categorize the products based on the shelves they are in. A shelf is a group of products presented together on the same e-commerce webpage, which usually contains products under the same categories \cite{tom}. Since our dataset does not contain the webpage information necessary to form shelves, our method was to remove the classes containing less than 400 products.

\begin{figure*}[!hbt]
    \centering
    \includegraphics[width=0.8\textwidth]{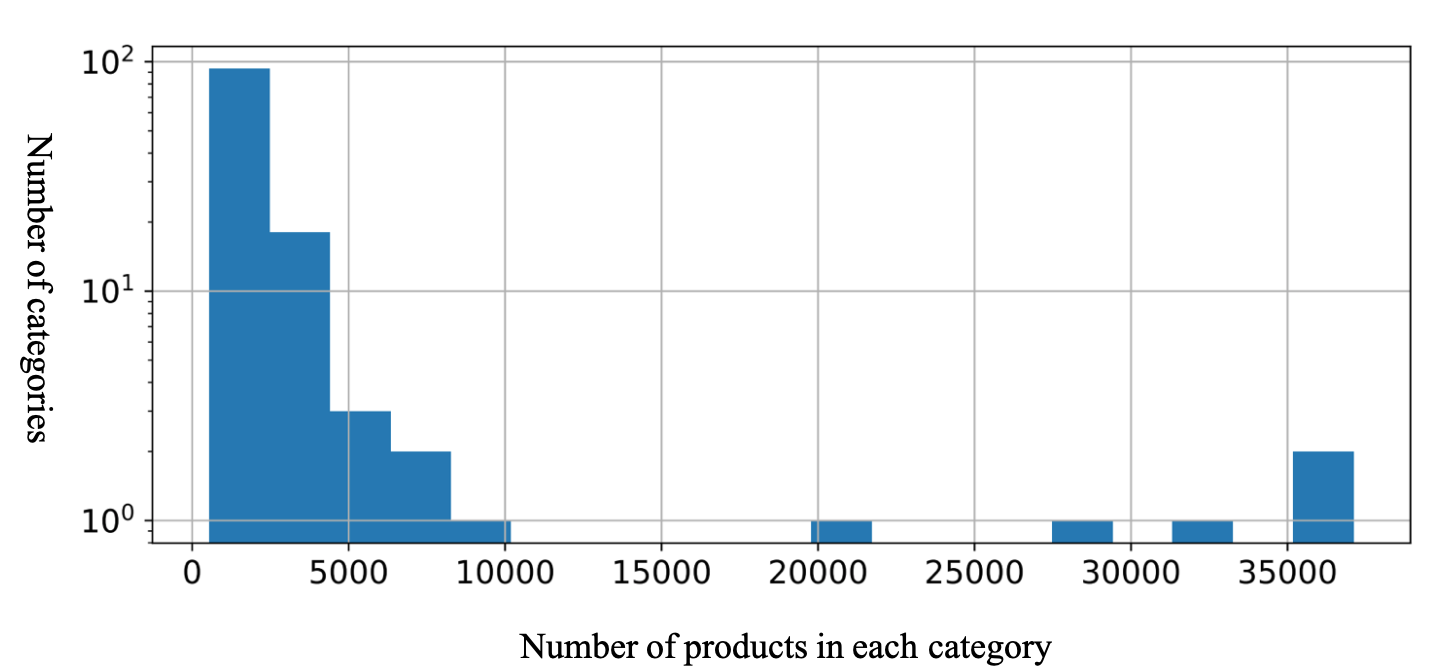}
    \caption{The $x$-axis represents the number of products in a category, whereas the $y$-axis represents the number of categories with that number of products.}
    \label{fig:2}
\end{figure*}

On average, each product belongs to 3 categories after pre-processing. The maximum number of products in a category is 37,102 and the minimum number of products in a category is 558. On average, there are 2,919 products per category. In addition, we can see from Fig.~\ref{fig:2} that the number of products per categories is not evenly distributed, which could introduce bias into the model.

\section{Baseline Models}
In order to understand how much we benefit from fusing the different modal classifiers, we report the baseline results for each modal below. We evaluate our results using the $F_1$ score (micro-averaged), which is an accepted metric for multi-label classification and imbalanced datasets \cite{Yang}. During training, for all classifiers, we used Adam \cite{adam} as our optimizer and categorical cross-entropy as our loss function. To accommodate multi-labeling, the final activations for each classifier utilizes the sigmoid function. Although both titles and descriptions are textual data, we leverage their different use-cases by treating them as different modalities, allowing us to perform different pre-processing steps as described below. 

\subsection{Description Classifier}
The description was pre-processed to remove stop words, excessive whitespace, digits, punctuations, and words longer than 30 characters. In addition, sentences were truncated to 300 words. To classify the pre-processed descriptions, we slightly modified Kim's CNN architecture for sentence classification. Kim's architecture is a CNN with one layer of convolution on top of word vectors initialized using Word2Vec \cite{kim,mikolov2013efficient}. Max-pooling over time is then applied \cite{collobert}, which serves to capture the most important features. Finally, dropout is employed in the penultimate layer.

Unlike the models used in prior studies mentioned above \cite{kim}, we used GloVe as our embedding. Words not covered by GloVe were initialized randomly. For our dataset, GloVe covers only 61.0\% of the vocabulary from the description. Our first convolution layer uses a kernel of size 5 with 200 filters. We then performed global max pooling, followed by a fully connected layer of 170 units with ReLU activations. Our final layer is another densely connected layer of 122 units with sigmoid activation. This model achieves 77.0\% on the test set.

\subsection{Title Classifier}
Although an identical classifier to the description classifier was used for the title, the title data was pre-processed differently. For the title, we did not remove the stop words and limited or padded the text to 57 words. We again chose GloVe for the embedding, in which words not covered were initialized randomly. GloVe covers 77.0\% of the vocabulary from the title. This model achieves 82.7\% on the test set.

\subsection{Image Classifier}
We modified the ResNet-50 architecture \cite{he} from Keras by removing the final densely connected layer and adding a densely connected layer with 122 units to match the number of labels we have. In addition, we changed the final activation to be sigmoidal. ResNet-50 is based on the architecture that achieves competitive results compared to other state-of-the-art models for image classification. We also used the pre-trained imagenet weights, which has been trained on the imagenet dataset \cite{imagenet_cvpr09}, containing more than 14 million images. We kept the weights of the earlier layers frozen and trained only the deeper/later layers \cite{DBLP:journals/corr/YosinskiCBL14}. We experimented with the number of trainable layers, in which our top model was trained on the last 40 layers, achieving a $F_1$ score 61\% on the test set.

\subsection{Summary}

The results summarized in Table~\ref{table:1} underscore that the classifiers differ in discriminative powers as the title and description classifiers significantly outperform the image classifier. This result is consistent with previous similar studies reporting a significant difference between the image and title classifiers \cite{tom}. Moreover, we have shown that the description classifier also significantly outperforms the image classifier. Such results suggests that text can provide more relevant information regarding a product's categories. 

\vspace{-1em}
\begin{table}[!hbt]
    \centering
    \caption{$F_1$ scores for each individual classifier.}
    \smallskip
    \begin{tabular}{c c}
    \hline
    \textbf{Modal} & $\mathbf{F_1}$ \textbf{(\%)}\\ 
    \hline
    
    Image & $61.0$ \\ \hline
    Title  & $82.7$\\ \hline
    Description & $77.0$\\ \hline
    
    \end{tabular}
    \vspace{0.3em}
    \label{table:1}
\end{table}

\section{Error Analysis}

\begin{table*}[t]
\centering
\caption{The top 15 most misclassified categories/classes for each classifier. The fraction represents the number of products which should be predicted as class $c_i$, but is not, over the total number of products that is in $c_i$.}
\smallskip
\resizebox{0.95\textwidth}{!}{
\begin{tabular}{l|l|l}
    \hline
    \textbf{Image} & \textbf{Description} & \textbf{Title}\\ 
       \hline
       Martial Arts $(146/154)$ & Women $(107/134)$ & Horses $(167/265)$ \\ \hline
Ballpoint Pens $(192/203)$ & Accessories $(663/924)$ & Novelty, Costumes \& More $(176/303)$ \\ \hline
Reptiles \& Amphibians $(136/144)$ & Clothing, Shoes \& Jewelry $(481/686)$ & Women $(77/134)$ \\ \hline
Small Animals $(327/349)$ & Boating $(222/327)$ & Accessories $(521/924)$ \\ \hline
Chew Toys $(123/132)$ & Novelty, Costumes \& More $(194/303)$ & Feeding $(137/244)$ \\ \hline
Squeak Toys $(202/223)$ & Parts \& Components $(135/212)$ & Clothing, Shoes \& Jewelry $(384/686)$ \\ \hline
Cards \& Card Stock $(222/246)$ & Men $(183/291)$ & Hunting \& Tactical Knives $(94/175)$ \\ \hline
Filter Accessories $(108/120)$ & Chew Toys $(83/132)$ & Balls $(99/185)$ \\ \hline
Other Sports $(188/210)$ & Balls $(116/185)$ & Hunting Knives $(80/152)$ \\ \hline
Bedding $(154/174)$ & Boating \& Water Sports $(368/591)$ & Boating $(171/327)$ \\ \hline
Tape, Adhesives \& Fasteners $(231/262)$ & Tape, Adhesives \& Fasteners $(161/262)$ & Small Animals $(176/349)$ \\ \hline
Birds $(432/490)$ & Office Furniture \& Lighting $(340/556)$ & Men $(146/291)$ \\ \hline
Pumps \& Filters $(270/308)$ & Forms, Recordkeeping \& Money Handling $(196/323)$ & Chew Toys $(65/132)$ \\ \hline
Cages \& Accessories $(125/144)$ & Hunting Knives $(89/152)$ & Boating \& Water Sports $(275/591)$ \\ \hline
Horses $(229/265)$ & Team Sports $(228/399)$ & Carriers \& Travel Products $(65/142)$ \\ \hline
\end{tabular}
}
\label{table:2}
\end{table*}

Table~\ref{table:2} exhibits that the top misclassified categories for each classifier generally reflect their inadequate representation in the dataset. Recall that the average number of products per category is 2,919. The Accessories category contains the most products (924) out of all the misclassified categories, but it is still far below the average. In addition, we can see that the top misclassified categories for each classifier seldom overlap between the modal classifiers. For the categories that the image classifier performed poorly, the description and title classifiers perform better and vice versa. This suggests that we should be able to combine the classifiers to effectively complement each other's shortcomings for a more effective overall result.

\section{Multi-Modality}
As prior studies found that late fusion models were more effective than early fusion models \cite{aberg,tom}, here we focus our studies on improving late fusion. 

\subsection{Predefined Policies}
First we evaluated the efficacy of our models using predefined rules in order to compare with other non-static policies. We experimented with max policy and mean policy of the output from each of the classifiers. The max policy selects the highest output for each class prediction from among the image, label, and title classifiers. This can be represented as
\begin{equation} \label{eq:1}
    o_{\max} = \max(o_{\mathrm{image}},o_{\mathrm{title}},o_{\mathrm{description}}),
\end{equation}
where $o_{\mathrm{image}}, o_{\mathrm{title}},o_{\mathrm{description}} \in R^{122}$ represent the output from each classifier, and the mean policy can be represented as
\begin{equation}\label{eq:2}
    o_{\mathrm{mean}} = \frac{o_{\mathrm{image}} + o_{\mathrm{title}} + o_{\mathrm{description}}}{3}.
\end{equation}
Both mean and max policy resulted in lower $F_1$ scores when compared to the top classifier, which is the title classifier. The mean policy yielded 81.7\%, while the max yielded 78.8\% $F_1$ scores. Intuitively, each classifier contributes equally to the mean policy. Therefore, we would expect that the average performance is less than that of the best performer. For the max policy, the erroneous maximal outputs from the low performing classifiers detriment the ultimate predictions.

\subsection{Linear Regression}
We trained a simple ridge linear regression model to fuse the individual classifiers into a single classifier. The model achieves 83.0\% on the test set. The model can be considered as minimizing the mean-squared error loss function according to
\begin{equation}\label{eq:3}
        \min_{\boldsymbol{w}} || \boldsymbol{wX} - \boldsymbol{y}||_2^2 + \alpha ||\boldsymbol{w}||_2^2,
\end{equation}
where $\boldsymbol{y}$ is the true label, $\boldsymbol{wX}$ is the predicted label, and $\alpha$ is the $L_2$ regularization strength. Nevertheless, the simple non-static policy can outperform static policies above.

\subsection{Bi-Modal Fusion}
Prior studies involved two neural networks, one for classifying images and another for classifying titles, using late fusion \cite{tom}. For comparison purposes, we examined models developed from fusing two of the three modal networks in this study. The first fused network includes the image classifier's output and the title classifier's output in a similar fashion with prior studies \cite{tom}. We then fused the title classifier's output and description classifier's output for the second fused network and fused the image classifier's output and description classifier's output for the third network. All three networks were fused the same way, using a three layer neural network to concatenate the outputs from each of the classifiers. The first, second, and third layers contained 200, 150 and 122 units, respectively. All the activations were sigmoidal. The image-description, image-title, and description-title fused networks yielded $F_1$ scores of 82.0\%, 85.0\%, and 87.0\%, respectively (see Table~\ref{table:3}).

\subsection{Tri-Modal Fusion}

\begin{figure*}[t]
    \centering
    \includegraphics[width=0.8\textwidth]{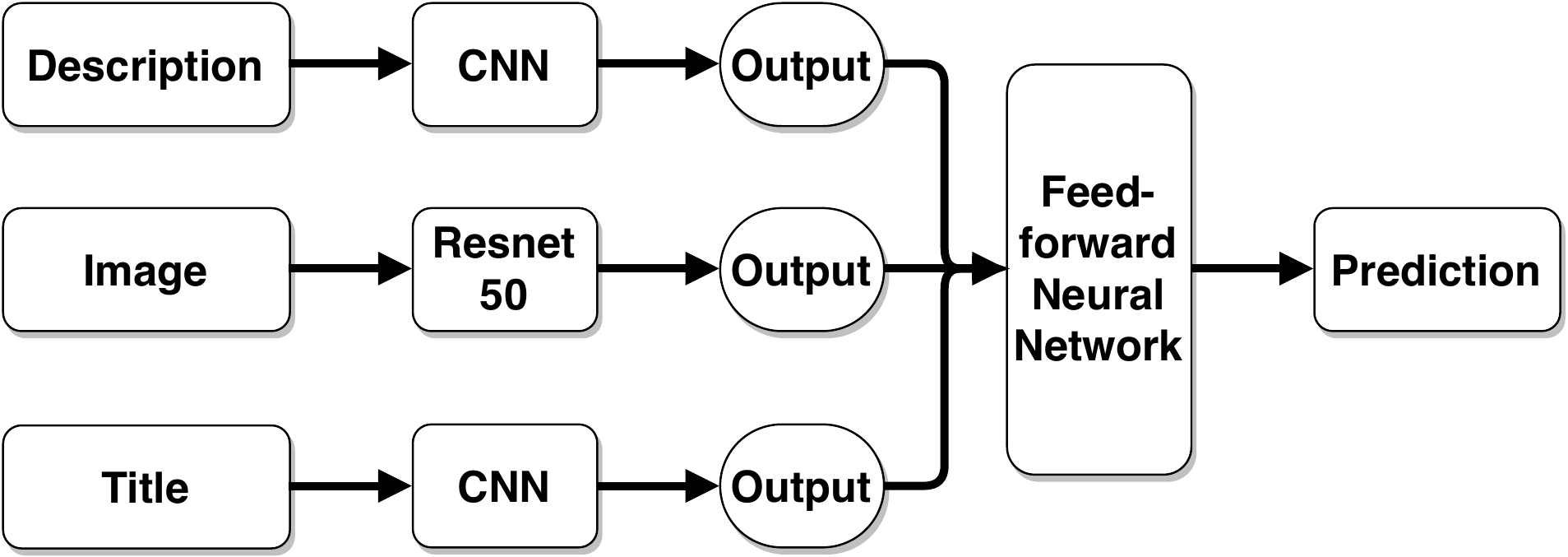}
    \caption{The proposed tri-modal fusion model architecture for product categorization using title, description and image.}
    \label{fig:3}
\end{figure*}

 Finally, we developed a tri-modal model to include the titles, images, and descriptions (see Fig.~\ref{fig:3}). To our knowledge, we are the first to fuse three classifiers/neural networks to categorize products. We fused the three classifiers as in the baseline models using a policy network, which is an additional neural network that takes in the output of each of the classifiers. We varied the number of layers, activation functions, and units of the neural networks. Through hyperparameter optimization, we found that the top policy network consists of three layers. It uses the sigmoidal activation on the first and last layers and hyperbolic tangent activation on the middle layer. This fused model achieves an $F_1$ score of 88.2\% surpassing all of the previous methods.

 \begin{table}[!hbt]
    \centering
        \caption{$F_1$ scores for fused classifiers.}
        \smallskip
    \begin{tabular}{c c}
    \hline
    \textbf{Model} & $\mathbf{F_1}$ \textbf{(\%)}\\ 
    \hline
    
    Max  & $78.8$\\ \hline
    Mean & $81.7$ \\ \hline
    Linear Regression & 83.0 \\ \hline
    Image-Description Fused & $82.0$\\ \hline
    Image-Title Fused & $85.0$\\ \hline
    Title-Description Fused & $87.0$\\ \hline
    Image-Description-Title Fused & $88.2$\\ \hline
    
    \end{tabular}
    \vspace{0.3em}

    \label{table:3}
\end{table}

\subsection{Discussion}
 
Compared to Table~\ref{table:2}, the proportion of misclassified products has reduced significantly in Table~\ref{table:4}. In examining Accessories, Horses, Clothing, and Shoes \& Jewelry, we can see that the proposed method outperforms the individual classifiers by a considerable margin. However, the proposed method fails to significantly reduce the number of misclassified products on certain categories, such as Chew Toys. According to Table~\ref{table:2}, each of the individual classifiers performed poorly predicting products as Chew Toys. This suggests that there remains categories that are underserved across all classifiers. To address this shortcoming, more data or other modes could be considered in future work. On the other hand, the result also suggests that as long as one classifier performs well on some of the tasks, it is sufficient for the overall model. For example, the number of misclassified products in Clothing, Shoes \& Jewelry dropped from 384 to 256. Overall, this method improves over the top individual classifier and the top two-modals fused network by 5.5\% and 1.2\%, respectively (see Table~\ref{table:3}).

\begin{table}
    \centering
    \caption{The top 15 most misclassified categories using our tri-modal fusion method.}
    \smallskip
     \begin{tabular}{c c}
    \hline
    & \textbf{Fused Image-Title-Description}\\ 
    \hline
1 & Chew Toys $(64/132)$ \\ \hline
2 & Accessories $(417/924)$\\ \hline
3 & Women $(58/134)$ \\ \hline
4 & Novelty, Costumes \& More $(127/303)$ \\ \hline
5 & Snacks $(143/363)$ \\ \hline
6 & Hunting Knives $(59/152)$ \\ \hline
7 & Men $(112/291)$\\ \hline
8 & Clothing, Shoes \& Jewelry $(256/686)$ \\ \hline
9 & Hunting \& Tactical Knives $(65/175)$\\ \hline
10 & Shampoos $(58/158)$\\ \hline
11 & Balls $(67/185)$\\ \hline
12 & Squeak Toys $(79/223)$ \\ \hline
13 & Horses $(92/265)$ \\ \hline
14 & Boating $(113/327)$ \\ \hline
15 & Airsoft $(68/200)$ \\ \hline
    \end{tabular}
    \vspace{0.3em}
\label{table:4}
\end{table}

\section{Conclusion}
We have shown that the title classifier can outperform the description classifier, and that the description classifier can outperform the image classifier. Moreover, a tri-modal fused network comprising of all three modalities outperformed any of the bi-modal fused networks. The performance improvements can be attributed to each of the classifiers addressing at least complementary portions of the tasks to account for the shortcomings of each individual classifier. 

While this study focused on late fusion, an early fusion approach can be explored in the future. In addition, more products, including products that may not fall under the predefined categories, can be added to reduce overfitting. A better text classifier can be built with pre-trained bi-directional contextualized language models \cite{devlin2018bert}. Transformers can be considered to replace CNNs and RNNs for both text and images \cite{Vaswani,artitw,Parmar}.

Finally, one possible extension to our work could be to build a vector representation of the products. Just as how word embeddings enabled us to more effectively classify text, a product embedding can be useful for capturing the relationship between products. Such a product embedding could help discover products that are ``similar'' for recommendation purposes and be used as input to a model to predict categories.

\bibliographystyle{aaai}
\bibliography{citation.bib}

\end{document}